# Enhancing Pavement Crack Classification with Bidirectional Cascaded Neural Networks

Taqwa I. Alhadidi
*Civil Engineering Department*
*Al-Ahliyya Amman University*
Amman, Jordan
t.alhadidi@ammanu.edu.jo

Asmaa Alazmi
*Department of Construction Project*
*Ministry of Public Work of Kuwait*
Kuwait City, Kuwait
dr_asmaa_alazmi@mpw.gov.kw

Shadi Jaradat
*CARRS-Q Centre for Data Science*
*Queensland University of Technology*
Brisbane, Australia
shadi.jaradat@hdr.qut.edu.au

Huthaifa I. Ashqar
*AI and Data Sceince Department*
*Arab American University*
Ramallah, Palestine
huthaifa.ashqar@aaup.edu

Ahmad Jaber
*Department of Transport Technology
and Economics Budapest University of
Technology and Economics,
Műegyetem rkp. 3., H-1111*
Budapest, Hungary
ahjaber6@edu.bme.hu

Mohammed Elhenawy
*CARRS-Q Centre for Data Science*
*Queensland University of Technology*
Brisbane, Australia
mohammed.elhenawy@qut.edu.au

*Abstract*— Pavement distress, such as cracks and potholes, is a significant issue affecting road safety and maintenance. In this study, we present the implementation and evaluation of Bidirectional Cascaded Neural Networks (BCNNs) for the classification of pavement crack images following image augmentation. We classified pavement cracks into three main categories: linear cracks, potholes, and fatigue cracks on an enhanced dataset utilizing U-Net 50 for image augmentation. The augmented dataset comprised 599 images. Our proposed BCNN model was designed to leverage both forward and backward information flows, with detection accuracy enhanced by its cascaded structure wherein each layer progressively refines the output of the preceding one. Our model achieved an overall accuracy of 87%, with precision, recall, and F1-score measures indicating high effectiveness across the categories. For fatigue cracks, the model recorded a precision of 0.87, recall of 0.83, and F1-score of 0.85 on 205 images. Linear cracks were detected with a precision of 0.81, recall of 0.89, and F1-score of 0.85 on 205 images, and potholes with a precision of 0.96, recall of 0.90, and F1-score of 0.93 on 189 images. The macro and weighted average of precision, recall, and F1-score were identical at 0.88, confirming the BCNN's excellent performance in classifying complex pavement crack patterns. This research demonstrates the potential of BCNNs to significantly enhance the accuracy and reliability of pavement distress classification, resulting in more effective and efficient pavement maintenance and management systems.

## I. INTRODUCTION

Pavement distress, including cracks and potholes, significantly challenges road safety [1] and maintenance[2]. These distresses compromise roadway integrity, increase accident risk, and lead to higher maintenance costs and reduced serviceability [3,4]. These distresses significantly affect ride quality and safety, leading to increased vehicle operating costs and potential accidents [5]. Various methodologies have been developed to identify and classify pavement distresses, including automated systems using machine learning [6–12] and deep learning[13–15]. Engineering research has made substantial use of machine learning (ML) approaches to improve predicted accuracy in a variety of fields. Applications of machine learning have proved essential to decision-making and process optimization[16]. Recent advancements in computer vision and artificial intelligence have enabled systems to accurately detect cracks and potholes from inspection images[17–20]. These systems enhance assessment efficiency and reduce reliance on labor-intensive manual inspections prone to human error [21].

A primary challenge in pavement crack classification is the diversity of crack types, which can appear visually similar. For instance, fatigue cracking can manifest as alligator (bottom-up) cracking or longitudinal (top-down) cracking, making differentiation difficult without sophisticated tools[22] . Automated systems for crack detection have gained traction, as they can analyze large datasets more efficiently than manual inspections. Liu et al. developed a deep convolutional neural network (CNN) that can detect and classify pavement cracks in complex backgrounds[3]. Machine learning techniques for pavement distress classification have shown promising results. Studies have demonstrated the potential of artificial neural networks (ANNs) and support vector machines (SVMs) in accurately predicting pavement conditions based on historical data [23–25]. These models can analyze patterns in large datasets, informing maintenance interventions. Integrating predictive analytics into pavement management systems can enhance maintenance efficiency and improve road safety. Pavement crack classification can be approached using various methodologies, including point cloud data from mobile laser scanning (MLS) systems. Studies highlight the importance of geometric features extracted from MLS data for effective crack detection and classification [26–30]. This method allows comprehensive pavement surface analysis, providing valuable information on crack morphology and distribution. Integrating deep learning techniques with MLS data has shown significant improvements in crack detection accuracy compared to traditional methods [31].

Accurate crack classification is crucial for long-term pavement performance assessments. The International Roughness Index (IRI) is a critical measure for evaluating pavement conditions, and accurate crack classification contributes to more reliable IRI predictions[32]. Understanding the relationship between crack types and overall pavement performance allows transportation agencies to develop more effective maintenance strategies prioritizing high-risk areas. Despite the significant advancements in machine learning (ML) techniques, the challenge of accurately classifying various types of pavement cracks remains a pressing issue in civil engineering. The complexity of pavement distress manifests in numerous forms, including top-down, bottom-up, and reflective cracking, which often exhibit visually similar patterns, complicating their identification and classification in real-world scenarios [22,33]. This difficulty is exacerbated by the limitations of existing ML models, which struggle to generalize across different datasets and environmental conditions, leading to inconsistent performance in crack detection tasks[31,34].

Bidirectional cascade neural networks (BCNNs) represent a significant advancement in the field of artificial intelligence, particularly in the domain of deep learning. These networks are characterized by their ability to process information in both forward and backward directions, which enhances their capacity to capture temporal dependencies in sequential data. The architecture of BCNNs typically integrates elements from both bidirectional recurrent neural networks (RNNs) and cascade neural networks, allowing for a more nuanced understanding of complex data patterns. The cascading nature of BCNNs further enhances their performance by enabling the model to refine its predictions through successive layers. Each layer can focus on different aspects of the input data, allowing for a more comprehensive analysis. For example, in the context of image processing, cascading convolutional neural networks (CNNs) can be employed to detect features at multiple scales, improving the robustness of the model against variations in input[35,36].

This paper presents significant advancements in pavement distress analysis through the application of Bidirectional Cascaded Neural Networks (BCNNs). We introduce the innovative use of BCNNs for classifying diverse pavement distress types and enhancing image processing with U-Net 50 augmentation. The use of Bidirectional Cascaded Neural Networks (BCNNs) for pavement distress classification is a game-changer for transportation infrastructure management. By utilizing BCNNs to interpret complex visual data, agencies can dramatically improve the precision of pavement assessments. This shift in paradigm to an automated, data-based approach enhances the detection and classification of pavement distresses with minimal reliance on manual inspection, reducing time and cost in road maintenance. BCNNs not only detect issues but also enhance knowledge of pavement conditions, facilitating predictive maintenance that prevents widespread damage. Their strategic use in pavement management optimizes maintenance and allows for safer, more reliable roads.

The organization of the paper is as follows: Section II provides a review of the literature on AI in pavement distress analysis. Section III explains the suggested framework, including the use of Bidirectional Cascaded Neural Networks (BCNNs) for the classification of distress types, such as system architecture, data sources, and processing procedures. Section IV presents case studies of the practical application of the BCNN framework, with strategic locations like intersections being highlighted. Section V presents summaries of the contributions and principal conclusions of the research, suggesting means to further enhance the BCNN framework through enhanced AI incorporation and extended spaces of application.

## II. LITERATURE REVIEW

Recent studies have explored a variety of deep learning architectures to enhance the classification accuracy of pavement cracks. For instance, Khanapuri et al. employed multiple deep learning methods, including Single Shot Detector (SSD) and Faster R-CNN, to analyze pavement damage, demonstrating the potential of these techniques in automated crack detection[31]. Similarly, Dhakal et al. developed artificial neural networks (ANN) and convolutional neural networks (CNN) to differentiate between types of pavement cracks, highlighting the effectiveness of these models in addressing the classification challenge[33]. However, despite these advancements, the models often require extensive training datasets and may not perform well under varying field conditions, indicating a need for further refinement and robust validation [33,34]

Despite the potential of these advanced techniques, the practical implementation of robust ML models for pavement crack classification remains hindered by several factors. One significant challenge is the need for large, diverse datasets that accurately represent the various types of cracks encountered in the field. Many existing studies rely on limited datasets, which can lead to overfitting and reduced generalizability of the models[31,33,34] .

In addition to data limitations, the interpretability of machine learning models poses another challenge in the context of pavement crack classification. Many deep learning models operate as "black boxes," making it difficult for engineers to understand the decision-making process behind the classifications[32] . Therefore, there is a growing need for research focused on enhancing the interpretability of ML models while maintaining their classification performance.

Furthermore, the integration of multi-modal data sources, such as combining visual data with sensor data, has been proposed as a means to improve classification accuracy. For instance, Yang et al. utilized ground-penetrating radar (GPR) alongside deep learning methods to detect structural distress in pavements, demonstrating that the fusion of different data types can enhance detection capabilities [37]. This approach highlights the potential for developing more comprehensive models that leverage diverse information sources to improve classification outcomes.

## III. METHODOLOGY

In this section, the proposed methodology is presented. The proposed methodology is shown in Fig.1.

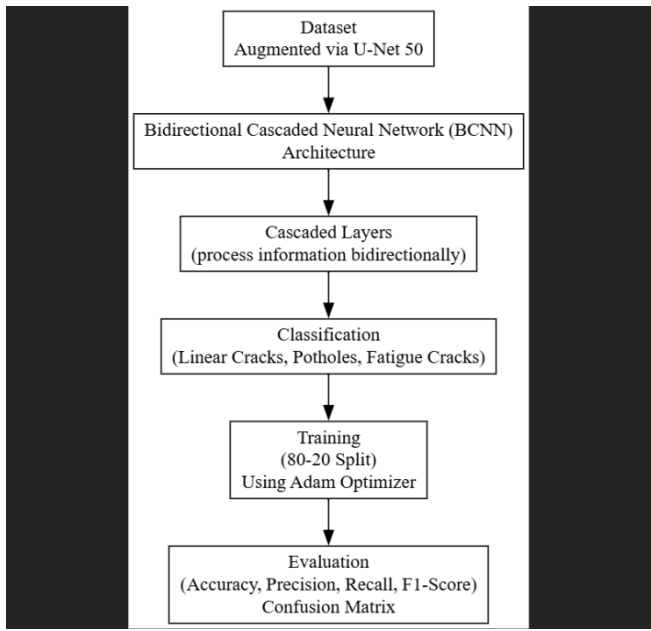

Fig. 1. Proposed Research Methodology

The flowchart describes the methodology using a Bidirectional Cascaded Neural Network (BCNN) to classify pavement crack images into three types: linear cracks, potholes, and fatigue cracks. Initially, the dataset of pavement images undergoes augmentation with U-Net 50, enhancing the quality and variability of the data. This dataset is then processed through a BCNN, which utilizes cascaded layers to refine features and improve accuracy by processing information bidirectionally. The classified images are then split into an 80-20 ratio for training and validation, employing the Adam optimizer to adjust learning rates efficiently. Finally, the model's effectiveness is evaluated using metrics like accuracy, precision, recall, and F1-score, alongside a confusion matrix to detail classification performance across categories. This structured approach ensures detailed scrutiny and optimization of the model's ability to accurately classify different types of pavement distress, aiming to enhance road maintenance and safety.

The primary objective of this study is to investigate the effectiveness of Bidirectional Cascaded Neural Networks (BCNNs) in classifying different types of pavement distress, specifically focusing on linear cracks, potholes, and fatigue cracks. This approach aims to enhance the accuracy and reliability of pavement condition assessments, crucial for maintaining road safety and infrastructure management.

*A.  Dataset and Image Augmentation*

The study utilizes a dataset comprising of 2455 images for training and 599 pavement images for testing dataset. categorized into three distinct classes: linear cracks, potholes, and fatigue cracks. To augment the dataset and improve the robustness of the BCNN, we employed the U-Net 50 model. A sample of the augmented dataset is shown in Fig.2. the dataset was collected from different online available resources and grouped by three main stresses types namely ; linear cracks ( longitudinal crackss, lateral cracks and diagonal cracks), fatigue cracks( i.e. blocks cracks, and alligator cracks), and potholes cracks. The trained dataset has 863 images for linear cracks, 822 fatigue cracks image, and 770 images for potholes images  This augmentation process involved generating variations of the original images through techniques such as rotation, scaling, and brightness adjustment, thereby creating a more comprehensive set for training the neural network.

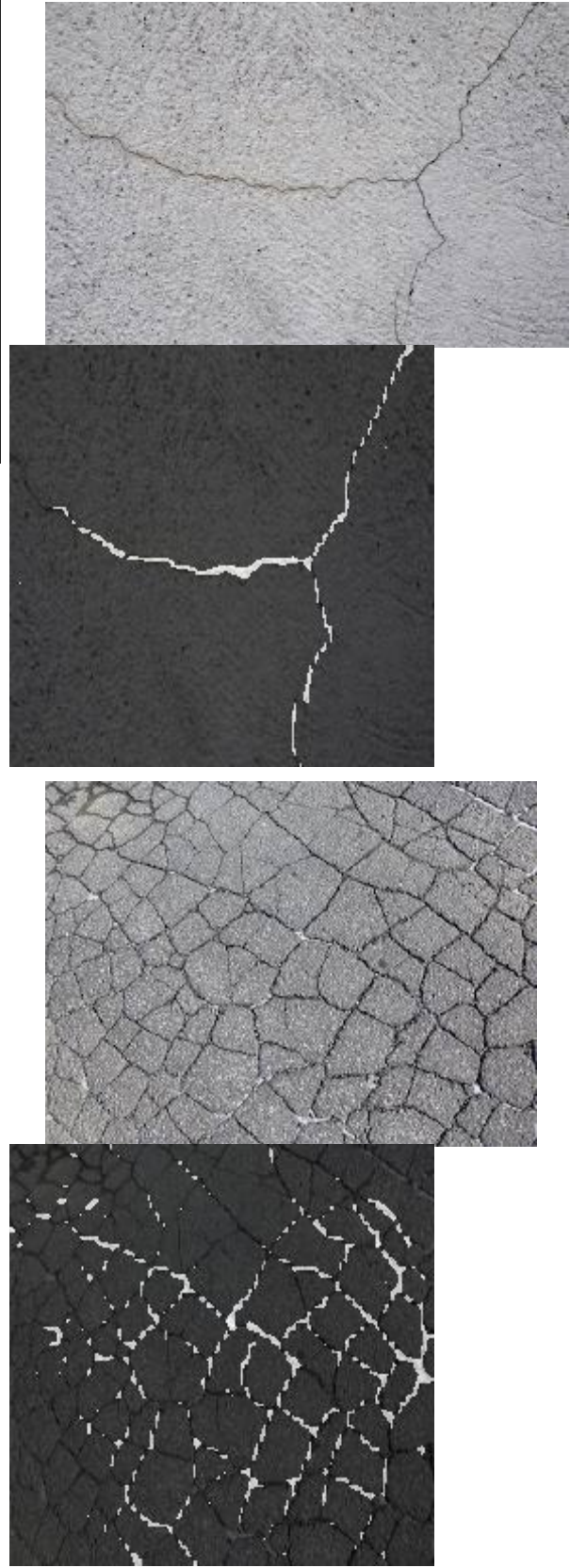

Fig. 2. Augmentation Results

## B. BCNN Architecture

The BCNN model implemented in this study is designed to process information bidirectionally, allowing it to capture temporal dynamics and contextual relationships more effectively than traditional single-direction neural networks. Each layer of the network is responsible for extracting and refining features from the input images, with the cascading structure ensuring that each successive layer builds upon the refined outputs of the previous one. This architecture is particularly suited to handle the complexities involved in distinguishing between similar types of pavement distress.

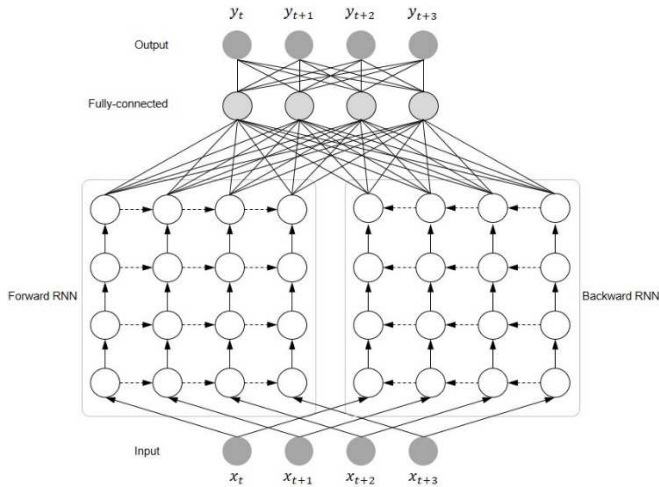

Fig. 3. BCNN Architecture:

## C. Training the BCNN

The BCNN was trained using a split of 75% of the images for training and 25% for validation. We employed the Adam optimizer for its efficient computation and adaptive learning rate capabilities, which helps in converging faster. The network was trained over multiple epochs, with real-time monitoring of performance metrics such as loss and accuracy on both training and validation sets to avoid overfitting.

## D. Evaluation Metrics

The performance of the BCNN was evaluated based on standard classification metrics: accuracy, precision, recall, and F1-score. These metrics provide a comprehensive understanding of the model's effectiveness across different classes and its ability to generalize across unseen data.

Our study employed a Bidirectional Cascaded Neural Network (BCNN) to classify pavement distress into three categories: fatigue cracks, linear cracks, and potholes. The model's performance was assessed based on its accuracy, precision, recall, and F1-score over a dataset of 599 images, which were augmented using the U-Net 50 model to enhance the model's ability to generalize across unseen data.

## IV. RESULTS

### A. Training Dynamics

The training and validation dynamics of the model are depicted in Fig 4. Throughout the training process, which spanned 15 epochs, both the accuracy and loss metrics for the training and validation sets were monitored.

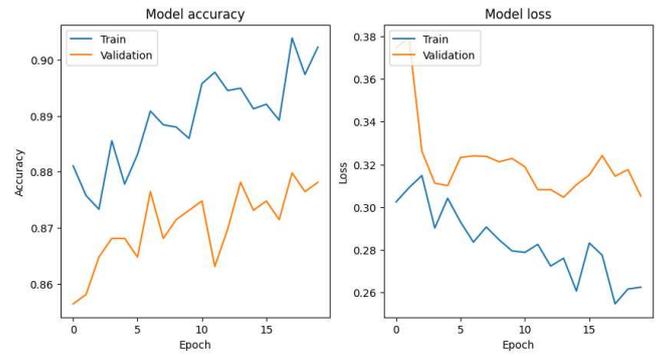

Fig. 4. Model Training and Validation Curves

The accuracy for the training set showed a general upward trend, indicating that the model was effectively learning the task. However, the validation accuracy displayed fluctuations, suggesting some challenges in the model generalizing to new data. Similarly, the training loss decreased steadily, reflecting the model's growing confidence in its predictions, while the validation loss exhibited variability, indicating potential overfitting.

### B. Classification Performance

The detailed performance of the model across the different classes of pavement distress is summarized in the table below:

TABLE I. CLASSIFICATION METRICS

| Class | Evaluation Metrics | | |
|---|---|---|---|
| | *Precision* | *Recall* | *F1-Score* |
| Fatigue cracks | 0.87 | 0.83 | 0.85 |
| Linear cracks | 0.81 | 0.89 | 0.85 |
| Potholes | 0.96 | 0.90 | 0.93 |

The model demonstrated high precision, especially in detecting potholes, with a precision of 0.96. The recall was highest for linear cracks at 0.89, indicating strong sensitivity in identifying this condition. The overall accuracy of the model was 0.87, with a weighted average F1-score of 0.88, showing balanced performance across precision and recall.

## V. DISCUSSION

The fluctuating validation accuracy and loss suggest that while the model is capable of high performance, improvements could be made to enhance its stability and generalization. Techniques such as introducing dropout, regularization, or more sophisticated data augmentation strategies might mitigate the observed overfitting.

These results underscore the BCNN's potential in classifying complex pavement crack patterns, which could significantly aid in the maintenance and management of transportation infrastructure. Further refinement and validation on larger, more varied datasets could help solidify these findings and support the deployment of such models in real-world applications.

## VI. CONCLUSION

This study demonstrated the potential of Bidirectional Cascaded Neural Networks (BCNNs) in accurately classifying different types of pavement distress, including fatigue cracks, linear cracks, and potholes. Through the application of a dataset enhanced by U-Net 50, the BCNN model achieved notable classification accuracy, demonstrating its efficacy in distinguishing between various pavement conditions critical for road safety and maintenance.

The model's overall accuracy was recorded at 87%, with precision and recall metrics indicating strong predictive performance across all categories. Notably, the model exhibited exceptional precision in detecting potholes (96%) and high recall for linear cracks (89%), which are often challenging to distinguish using traditional methods. The balanced F1-scores across categories further confirmed the model's robustness, making it a reliable tool for automated pavement inspection systems.

However, the variability observed in the validation accuracy and loss suggests that while the model performs well on known data, its ability to generalize to new, unseen datasets could be improved. Addressing this potential overfitting through advanced regularization techniques, enhanced data augmentation, or by refining the model architecture could lead to even more reliable and generalizable results.

The findings from this research not only highlight the capabilities of BCNNs in a practical application but also open avenues for further studies to explore the integration of such models into real-world transportation infrastructure management systems. By continuing to enhance the accuracy and reliability of these models, there is significant potential to reduce costs associated with road maintenance and improve overall traffic safety.

Future work should aim to validate these results on larger and more diverse datasets, possibly integrating real-time data to dynamically assess pavement conditions. Additionally, expanding the model to include other forms of pavement distress and exploring the integration with other smart transportation systems could provide comprehensive solutions for urban planning and smart city development.